\documentclass[10pt,twocolumn,letterpaper]{article}

\usepackage{cvpr}              

\usepackage{graphicx}
\usepackage{amsmath}
\usepackage{amssymb}
\usepackage{booktabs}
\usepackage{colortbl}

\usepackage{multirow}
\usepackage{subcaption}
\usepackage{tabularx}
\usepackage{lipsum}

\newcolumntype{Y}{>{\centering\arraybackslash}X}

\definecolor{cvprblue}{rgb}{0.21,0.49,0.74}
\usepackage[pagebackref,breaklinks,colorlinks,allcolors=cvprblue]{hyperref}

\def\paperID{*****} 
\def\confName{CVPR}
\def\confYear{2025}

\title{Towards Unconstrained 2D Pose Estimation of the Human Spine} 

\author{Muhammad Saif Ullah Khan \qquad Stephan Krauß \qquad Didier Stricker \\
German Research Center for Artificial Intelligence (DFKI) \\
Trippstadter Str. 122, 67663 Kaiserslautern Germany\\
{\tt\small \url{https://saifkhichi96.github.io/research/spinepose/}}
}

\begin{document}
\twocolumn[{%
\renewcommand\twocolumn[1][]{#1}%
\maketitle
\includegraphics[clip,trim={0 1cm 0 0.6cm},width=\linewidth]{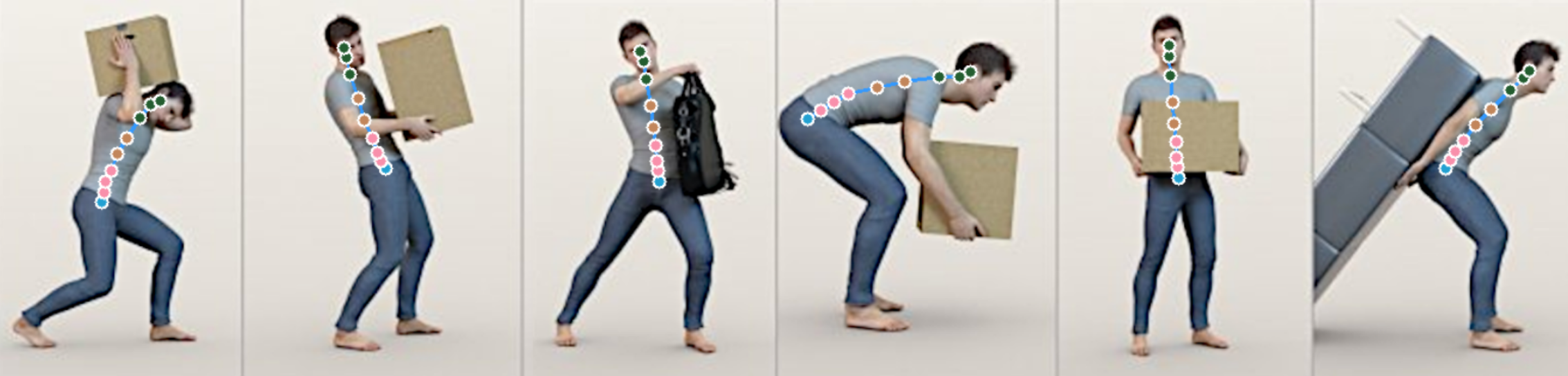}
\label{fig:teaser}
}]

\begin{abstract}
We present \textbf{SpineTrack}, the first comprehensive dataset for 2D spine pose estimation in unconstrained settings, addressing a crucial need in sports analytics, healthcare, and realistic animation. Existing pose datasets often simplify the spine to a single rigid segment, overlooking the nuanced articulation required for accurate motion analysis. In contrast, SpineTrack annotates nine detailed spinal keypoints across two complementary subsets: a synthetic set comprising 25k annotations created using Unreal Engine with biomechanical alignment through OpenSim, and a real-world set comprising over 33k annotations curated via an active learning pipeline that iteratively refines automated annotations with human feedback. This integrated approach ensures anatomically consistent labels at scale, even for challenging, in-the-wild images. We further introduce \textbf{SpinePose}, extending state-of-the-art body pose estimators using knowledge distillation and an anatomical regularization strategy to jointly predict body and spine keypoints. Our experiments in both general and sports-specific contexts validate the effectiveness of SpineTrack for precise spine pose estimation, establishing a robust foundation for future research in advanced biomechanical analysis and 3D spine reconstruction in the wild.
\end{abstract}

\section{Introduction}
\label{sec:intro}

Human pose estimation (HPE) has advanced remarkably over the past decade, driven by deep learning and large-scale annotated datasets. Yet, most existing approaches and benchmarks—such as COCO~\cite{lin2014microsoft}, MPII~\cite{andriluka2014cvpr}, and Halpe~\cite{alphapose}—primarily focus on limb and facial keypoints. In these datasets, the spine is typically represented by a single rigid segment or a minimal set of markers (e.g., neck and pelvis), which fails to capture the curvature and dynamic articulation of the human spinal column. This simplified representation limits the utility of pose estimation systems in applications where detailed spine analysis is critical, such as sports performance assessment, injury prevention, rehabilitation, ergonomic evaluation, and realistic avatars.

Accurate tracking of spinal motion is especially crucial in sports science, where subtle variations in alignment can reveal an athlete's posture, balance, and injury risks. Activities such as weightlifting, diving, hockey, or archery all rely on precise spinal control, yet vertebral landmarks are not directly observable and must be inferred from surrounding anatomical cues, making annotation subjective and labor-intensive. A major obstacle to advancing spine pose estimation in the wild is therefore the lack of labeled data with fine-grained vertebral markers that can generalize beyond controlled settings. Although Halpe and MPII provide spine endpoints, and 3D datasets like Human3.6M~\cite{h36m_pami} and AMASS~\cite{mahmood2019amass} include some spine keypoints, none offer detailed annotations for unconstrained environments. This stems from the high cost of labeling large sets of real-world images, compounded by the difficulty of consistently identifying internal joints across diverse poses and anatomies.

To address these limitations, we present \textbf{SpineTrack}, the first dataset specifically designed for detailed 2D spine pose estimation. SpineTrack introduces a fine-grained representation of the spinal column by annotating nine vertebral keypoints, capturing the curvature and dynamic behavior of the spine across diverse motion states. Our dataset comprises a real-world set with high-fidelity spine annotations obtained through an active learning-based annotation process, and a synthetic set generated via an Unreal Engine-based pipeline with accurate ground-truth labels under controlled conditions. The real-world images reflect the variability of natural environments, including occlusions, lighting changes, and anatomical diversity, while the synthetic images bootstrap the annotation process and provide a rich source of training data for rare or challenging poses.

Complementing our dataset, we propose the \textbf{SpinePose} framework that builds upon traditional body pose models. SpinePose incorporates precise spine keypoints into the prediction head of a pre-trained body pose network using knowledge distillation. A targeted regularization approach maintains a balance between assimilating new spine annotations and retaining the existing body keypoint information in varied settings. A major feature of our framework is the integration of anatomical constraints into the optimization process using novel body structure and spine smoothing loss functions. Our contributions are summarized as follows:

\begin{itemize}
    \item We introduce \emph{SpineTrack}, a novel dataset for 2D spine pose estimation, comprising 50,962 images and 58,766 annotated humans across both real-world and synthetic images with body, feet, and spine annotations.
    \item We propose \emph{SpinePose}, a distillation-based framework with \emph{specialized loss functions} that seamlessly integrates detailed spine keypoints into existing body pose estimation architectures, achieving high accuracy across both standard benchmarks and sports-specific scenarios.
    \item We use a biomechanical validation framework that leverages OpenSim to assess and ensure the anatomical plausibility of spine annotations.
    \item Extensive experiments validate our approach, demonstrating significant advances in spine tracking without compromising overall pose estimation performance.
\end{itemize}


\section{Related Work}
\label{sec:related_work}

Accurate tracking of human spine motion is a longstanding challenge in biomechanics, sports science, and medical applications. Traditional motion capture systems rely on optical markers or inertial measurement units (IMUs), but these methods have limitations in real-world deployment. Recent advancements in markerless vision-based deep learning approaches have enabled full-body pose tracking, yet existing models lack explicit spinal articulation. This section reviews relevant work in spine pose estimation and highlights the gaps our work aims to bridge.

\paragraph{Traditional Spine Motion Tracking} While markerless spine motion estimation in the wild remains an open challenge, traditional methods including specialized hardware~\cite{kam2017low,bogo2017dynamic}, inertial sensors~\cite{miezal2014ambulatory,schepers2018xsens,hajibozorgi2016sagittal,voinea2016measurement}, and reflective markers~\cite{merriaux2017study,panero2021multi,tanie2005high} have been long used to track spinal motion in controlled settings. Optical systems (e.g., Vicon) remain the gold standard for spine tracking due to their sub-millimeter accuracy~\cite{merriaux2017study}. These systems track reflective markers placed along the spinous processes and torso. However, their use is restricted to controlled environments due to cost, setup complexity, and occlusion issues, particularly during extreme spinal movements~\cite{rast2016between,hachmann2023human}. IMU-based approaches offer a more portable alternative by placing gyroscopes and accelerometers along the spine to estimate orientation. Commercial systems achieve reasonable accuracy for coarse trunk movements but lack fine-grained vertebral tracking due to sensor drift and soft-tissue artifacts~\cite{schepers2018xsens}. More specialized multi-IMU configurations~\cite{hajibozorgi2016sagittal,voinea2016measurement} improve fidelity but remain cumbersome for daily use.

\paragraph{Markerless Human Pose and Spine Tracking} Deep learning has driven significant advances in markerless pose estimation from RGB and RGB-D inputs. Methods like OpenPose~\cite{cao2019openpose}, AlphaPose~\cite{alphapose}, and RTMPose~\cite{jiang2023rtmpose} achieve robust 2D pose estimation. This also includes a neck and hip center point, which can represent the spine as a single, rigid segment. To our knowledge, no existing models directly regress 2D keypoints on the spine from input images. Some 3D pose estimation works can directly regress the spine midpoint, but the fidelity remains low~\cite{h36m_pami}. Other recent 3D works fit parametric models like SMPL~\cite{loper2023smpl} to reconstruct full-body meshes, including the spinal regions. However, they still treat the spine as a low-flexibility segment. Multi-camera setups, such as OpenCap~\cite{uhlrich2023opencap} and Pose2Sim~\cite{pagnon2021pose2sim}, improve tracking accuracy by fusing multiple views with biomechanical models~\cite{seth2011opensim}, enabling near-lab precision for spino-pelvic motion. Inverse kinematics (IK) driven estimates of spinal coordinates can also be computed from the biomechanical model~\cite{pagnon2021pose2sim}. However, these require controlled capture environments, limiting generalization. Some commercial depth cameras, like Microsoft Kinect~\cite{Kean2011} and StereoLabs ZED 2i, can predict basic spine landmarks (e.g., base, mid). However, they provide a coarse representation and have errors in curvature estimation~\cite{castro2017evaluation}.
\begin{figure*}[ht]
    \centering
    \includegraphics[width=0.9\linewidth]{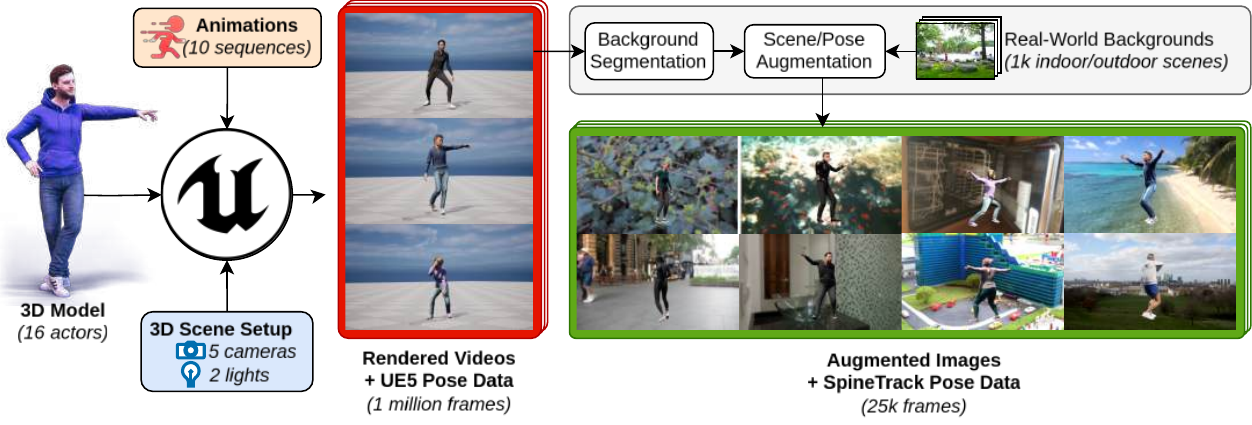}
    \caption{\textbf{SpineTrack-Unreal Generation}. We show the data generation pipeline with Unreal Engine~\cite{unrealengine} which we use to obtain synthetic images and ground-truth locations of keypoints roughly corresponding to SpineTrack skeleton. Keypoint positions are refined by scaling the biomechanical model to each actor and using an inverse kinematics solver to recompute marker positions. Images are augmented with diverse scenes from OpenImagesV7 dataset~\cite{OpenImages} using SAM~\cite{kirillov2023segment} to simulate more realistic backgrounds.}
    \label{fig:spinetrack-unreal}
\end{figure*}

\section{Methodology}
\label{sec:methodology}

We present the SpineTrack dataset in Sec.~\ref{sec:spinetrack} that captures detailed anatomical information on the human spine in unconfined settings. The annotations are created using an active learning pipeline~\cite{amin2023deep} where human annotators and a neural network support each other. We first explain the synthetic set, called \emph{SpineTrack-Unreal}, which is used to initialize the automated annotator in our pipeline. Generation process is illustrated in Fig.~\ref{fig:spinetrack-unreal}. The following subsection then describes \emph{SpineTrack-Real}, forming the primary component of our dataset. This section also describes our protocol for validating annotation correctness. Collectively, both sets form a robust foundation for training our SpinePose models in Sec.~\ref{sec:spinepose-framework}, which effectively integrate spine keypoints into their predictions while maintaining performance on benchmark in-the-wild datasets.

\subsection{SpineTrack Dataset}
\label{sec:spinetrack}

\begin{figure}[ht]
    \centering
    \includegraphics[width=0.7\linewidth]{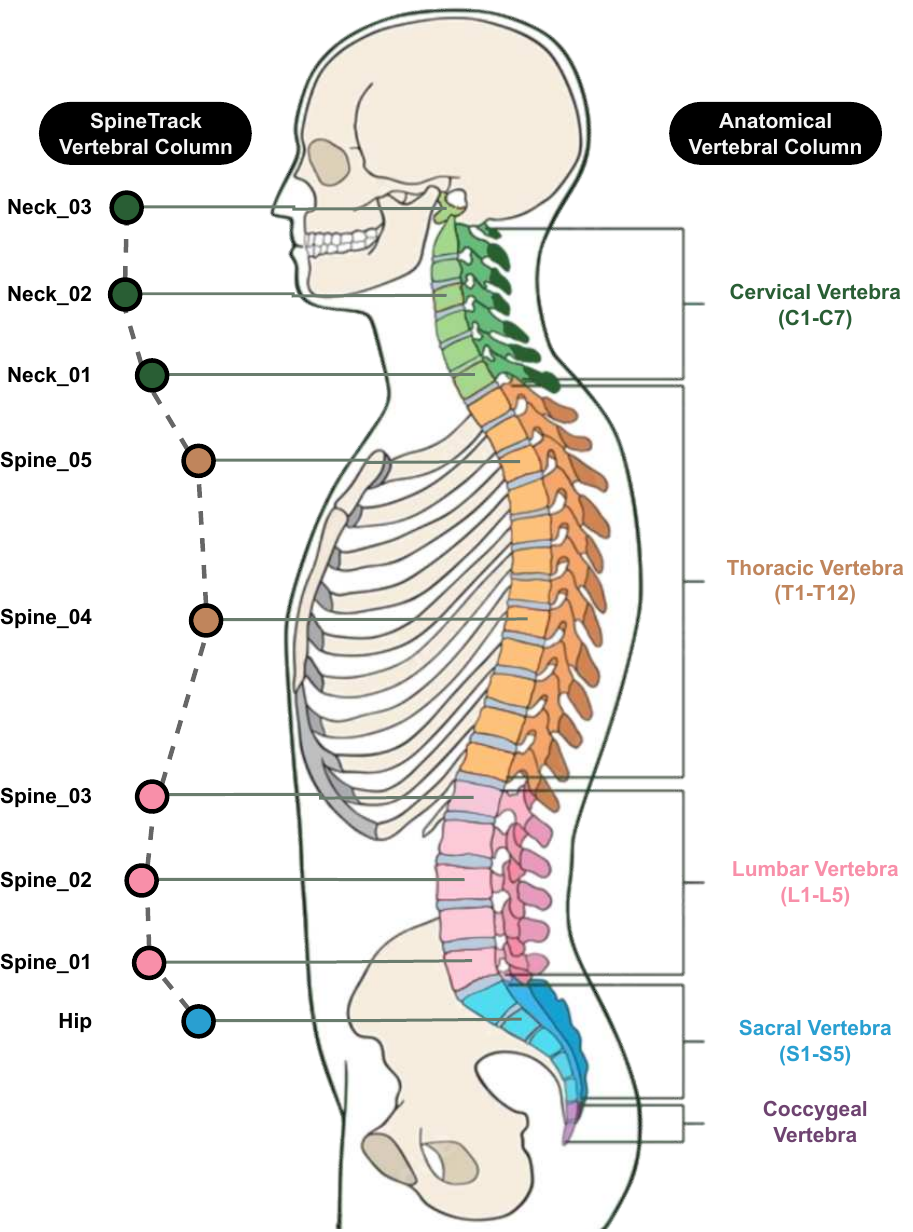}
    \caption{\textbf{Spine Keypoints.} We annotate nine keypoints along the spinal column to capture upper body movement without excessive overhead. Three are placed on the cervical spine (C1, C4, C7), two on the thoracic spine (T3, T8), three on the lumbar spine (L1, L3, L5), and one at the sacrum near the pelvis. This distribution balances anatomical realism and annotation cost, reflecting the distinct mobility of each spinal region.}
    \label{fig:skeleton-definition}
\end{figure}

\paragraph{Skeleton Configuration} 
In addition to standard body keypoints~\cite{lin2014microsoft}, we label \emph{nine vertebral landmarks} for refined modeling of upper body motion. The spinal column is structured into five regions, as shown in Fig.~\ref{fig:skeleton-definition}, of which three majorly contribute to upper body movements: the cervical, thoracic, and lumbar spines. In our design, three keypoints are assigned to the cervical spine to capture neck dynamics; five keypoints are placed along the thoracolumbar region to represent the trunk; and one keypoint marks the sacral region as the root joint. In addition, we also annotate the left and right sternoclavicular joints (not illustrated) to account for independent shoulder movement. Six keypoints on the feet are also provided, with a total of \emph{35 annotated keypoints}. This configuration is motivated by the need for accurate biomechanical modeling with downstream applications in healthcare and sports analysis in mind.

\subsubsection{SpineTrack-Unreal}
\label{sec:spinetrack-unreal}

The \emph{SpineTrack-Unreal} dataset was generated using Unreal Engine 5 with high-fidelity avatars. Sixteen diverse 3D human models were animated with ten distinct motion sequences—including walking, running, jumping, stretching, and twisting—and captured from five calibrated camera views under controlled lighting conditions. The UE5 skeleton system provided precise 3D annotations projected into 2D image space. To further enhance realism and diversity, the rendered images were blended with 1,000 real-world indoor and outdoor backgrounds obtained via the Segment Anything Model (SAM)~\cite{kirillov2023segment} from the OpenImagesV7 dataset~\cite{OpenImages}. This synthesis produced approximately 25k annotated frames covering various human motions and environmental conditions.

\begin{figure*}[ht]
    \centering
    \begin{minipage}[b]{0.17\linewidth}
        \includegraphics[width=\linewidth]{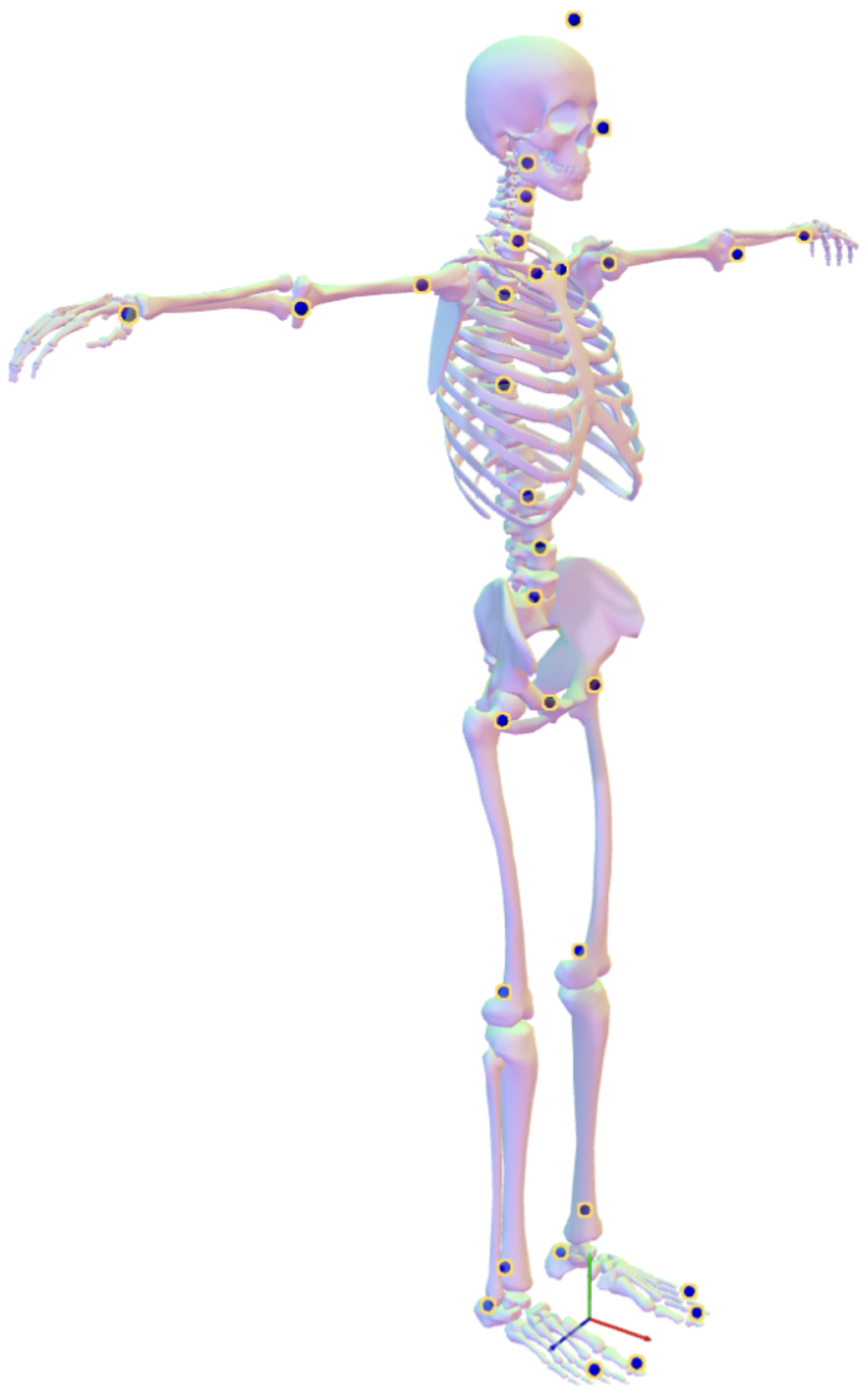}
    \end{minipage}
    \hfill
    \vrule width 1pt height 4.5cm
    \hfill
    \begin{minipage}[b]{0.75\linewidth}
        \includegraphics[width=\linewidth]{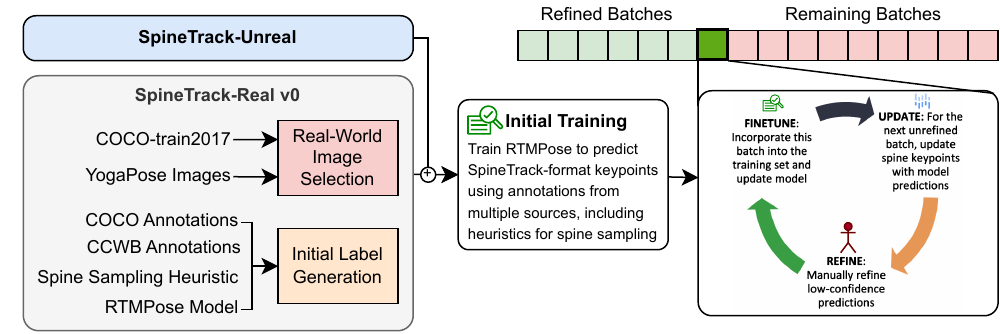}
    \end{minipage}%
    \caption{\textbf{SpineTrack-Real Creation.} On the \emph{left}, we show an OpenSim~\cite{seth2011opensim} model adapted from~\cite{rajagopal2016full,beaucage2019validation,pagnon2021pose2sim} with all 35 anatomical markers labeled in our dataset, including 17 COCO keypoints~\cite{lin2014microsoft}, head top, feet (six), spine (nine), and two sternoclavicular joints. On the \emph{right}, our iterative pipeline begins with real-world image selection and initial pseudo-labels generated by a pretrained model. A preliminary spine-aware model is trained on these labels plus SpineTrack-Unreal. Annotations are then refined in batches, where human annotators correct model predictions, and the improved labels are fed back into the training data to fine-tune the model. This process repeats for all batches, improving model accuracy and reducing data noise at each step.}
    \label{fig:spinetrack-real-annotation}
\end{figure*}

\subsubsection{SpineTrack-Real}
\label{sec:spinetrack-real}

While SpineTrack-Unreal provides precise synthetic annotations, it lacks the occlusions, lighting variations, and scene diversity found in real-world imagery. To bridge this gap, we construct SpineTrack-Real using an iterative pipeline (Fig.~\ref{fig:spinetrack-real-annotation}). We select images from multiple sources (e.g., COCO, YogaPose~\cite{pandit2020yoga}). Images are selected based on several criteria designed with annotation ease in mind (e.g., large people). Where ground-truth annotations are not available (e.g., feet, yoga images), we apply a pretrained model (RTMPose-L~\cite{jiang2023rtmpose}) to detect visible body and feet joints. Using shoulder and hip joints as drivers, we fit two smooth curves around the spinal region, forming a spine-like structure. Assuming constant inter-vertebral distance, initial spine guesstimates are sampled from these curves at predefined intervals. This results in over 33k real-world annotations and SpineTrack-format initial pseudo-labels.

\paragraph{Active Learning-Driven Annotation Refinement}

\begin{figure}[ht]
    \centering
    \includegraphics[width=\linewidth]{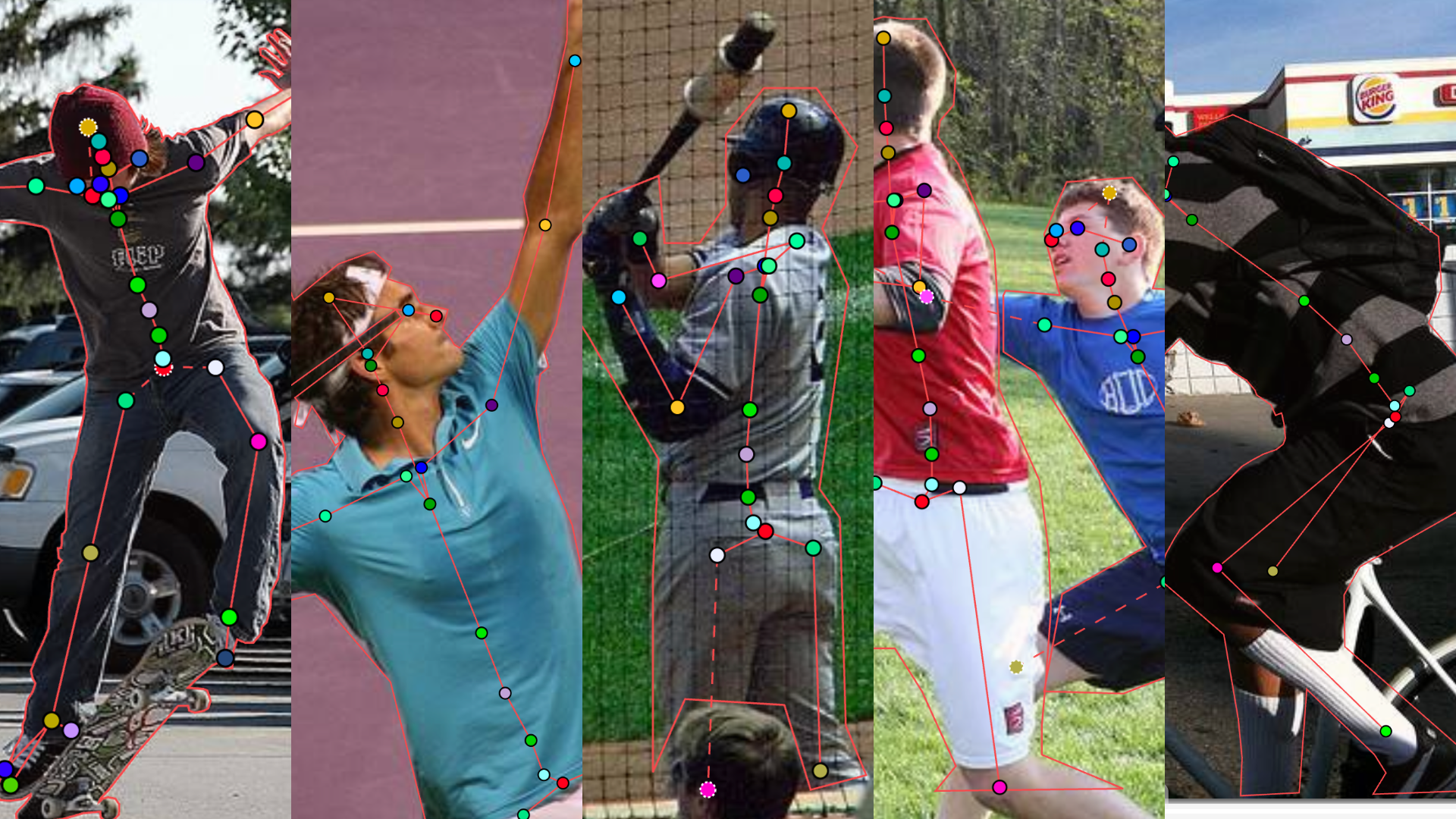}
    \caption{\textbf{SpineTrack Dataset.} Example images illustrating the range of human activities, occlusions, and body shapes in SpineTrack, with detailed spine keypoints and standard body landmarks.}
    \label{fig:spinetrack-annotations}
\end{figure}

These preliminary annotations are combined with SpineTrack-Unreal to train a \emph{spine-aware} pose model, establishing a baseline for in-the-wild spinal keypoint detection. Next, we split the dataset into batches and iteratively refine annotations. For each new batch, the trained model predicts spine keypoints; human annotators correct low-confidence or erroneous predictions, and the updated labels are merged back into the training set. The model is fine-tuned at each cycle, progressively improving annotations. This approach scales efficiently, requiring minimal human effort while capturing the anatomical complexity of the spine in real-world scenes. Fig~\ref{fig:spinetrack-annotations} shows samples from our real-world dataset with both body and detailed spine annotations.

\paragraph{Annotation Quality Validation}

We validate the anatomical correctness of our spine annotations through a biomechanical protocol with an OpenSim model fitted to motion capture sequences. This model is shown in Fig.~\ref{fig:spinetrack-real-annotation} (left). A calibrated multi-camera system records a controlled movement sequence, capturing a range of motions. We estimate keypoints from each camera view using the 2D pose model, then reconstruct the 3D pose via triangulation. Next, we scale the OpenSim model to the subject’s dimensions, and its inverse kinematics solver computes expected joint positions across all frames. Accuracy is measured by computing the root-mean-square error (RMSE) between the triangulated keypoints and these positions.

First, we evaluate a state-of-the-art pose model trained on diverse in-the-wild datasets, observing an RMSE of 2–4 cm for standard body keypoints. We use this range as a reference when evaluating SpineTrack annotations. A model trained only on SpineTrack-Real undergoes the same testing, and any keypoints showing an RMSE above 10 cm are manually refined in the annotation pipeline. We iterate until spine keypoints match or surpass the baseline’s accuracy, ensuring that SpineTrack annotations achieve a level of reliability comparable to conventional human pose datasets.

\subsection{SpinePose Architecture}
\label{sec:spinepose-framework}

\begin{figure}[ht]
    \centering
    \includegraphics[width=0.96\columnwidth,clip,trim={0 0 1cm 0}]{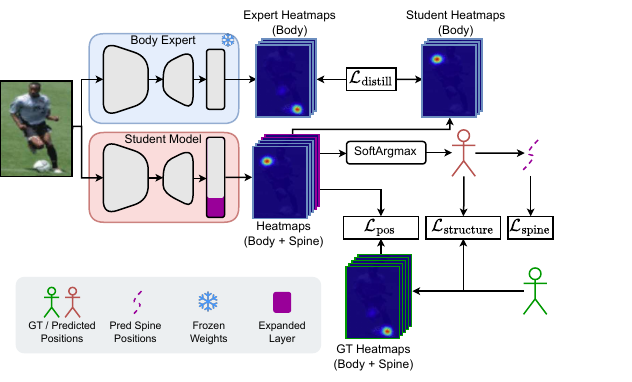}
    \caption{\textbf{SpinePose Architecture.} Our teacher–student approach for 2D spine pose estimation integrates knowledge from a pretrained body expert with newly introduced spine keypoints. The student model expands the teacher’s head to predict both body and spine heatmaps, using a combination of distillation losses, a positional keypoint loss, and our structure-based losses to achieve anatomically consistent predictions. The final objective balances accuracy on existing body joints with the newly added vertebral keypoints, ensuring robust performance in real-world scenarios.}
    \label{fig:spinepose-overview}
\end{figure}

We propose a distillation-based framework~\cite{hinton2015distilling} where we assume access to a body pose expert, $\mathcal{T}$, trained on a large-scale in-the-wild dataset. As illustrated in Fig.~\ref{fig:spinepose-overview}, we use this expert for teaching a student model, $\mathcal{S}$, to predict an \emph{extended} skeleton that includes both the teacher's body keypoints and additional spine keypoints annotated in our SpineTrack dataset. This is inspired by~\cite{khan2024continual} who propose a similar paradigm for retaining generalization when fine-tuning on smaller datasets. Nonetheless, our experiments in Sec.~\ref{sec:experiments} show that directly training on SpineTrack alone is sufficient to achieve competitive in-the-wild performance.

Following~\cite{khan2024continual}, we start with a pretrained teacher network and use it to spawn an identical student. The head architecture is modified by \emph{expanding the output layer(s)} to accommodate new spine keypoints. This involves increasing the channels in the layer(s) responsible for creating the final heatmaps. The student network is initialized with weights from the pretrained teacher, including the expanded layers where weights are only copied into original matrix rows. However, instead of filling new rows with zeros like~\cite{khan2024continual}, we initialize them from a normal distribution with statistics computed from the existing weights.

\paragraph{Problem Definition}

Let $\mathcal{K}_{\text{body}}$ be the set of body keypoints, and let $\mathcal{K}_{\text{spine}}$ be the additional keypoints introduced in SpineTrack. We define the extended skeleton as:
\begin{equation}
    \mathcal{K}_{\text{student}} = \mathcal{K}_{\text{body}} \cup \mathcal{K}_{\text{spine}},
\end{equation}
where $\mathcal{K}_{\text{body}} \cap \mathcal{K}_{\text{spine}} = \emptyset$ by construction. The student model $\mathcal{S}$ predicts heatmaps for all keypoints in $\mathcal{K}_{\text{student}}$, while the teacher model $\mathcal{T}$ only outputs heatmaps for $\mathcal{K}_{\text{body}}$. Given an input image $x$, let
\[
p_s = \mathcal{S}(x) \quad \text{and} \quad p_t = \mathcal{T}(x)
\]
denote the student's and teacher's predicted heatmaps, respectively. For each keypoint $k \in \mathcal{K}_{\text{body}}$, the teacher provides a probability distribution $p_{t,k}$ over pixel locations, and the student produces $p_{s,k}$ for the same keypoint. For $k \in \mathcal{K}_{\text{spine}}$, only the student outputs a distribution $p_{s,k}$, since the teacher lacks these keypoints. The ground-truth heatmap $g_k$ for each keypoint $k$ is obtained from our dataset.

We aim to train $\mathcal{S}$ such that it preserves the teacher's expertise on $\mathcal{K}_{\text{body}}$ through distillation, and accurately predicts $\mathcal{K}_{\text{spine}}$ for enhanced anatomical detail. 
This is achieved through the combination of loss functions introduced in the following section.

\subsubsection{Loss Functions}
\label{subsec:loss_functions}

We train the student model by combining four key objectives: a \emph{positional loss} for accurate keypoint localization, a \emph{distillation loss} to preserve the teacher’s knowledge, a \emph{structure loss} to ensure anatomical plausibility, and a smoothness loss to discourage sharp bends in the spine:

\paragraph{Positional (Keypoint) Loss}

The primary objective in standard pose estimation pipelines is to minimize the distance between predicted and ground-truth keypoint locations. Following~\cite{jiang2023rtmpose}, we use KL-divergence as the distance metric. Specifically, for each $k \in \mathcal{K}_{\text{student}}$, let $p_{s,k}$ and $g_k$ be the predicted and ground-truth distributions. We define
\begin{equation}
    \mathcal{L}_{\text{pos}} \;=\; \sum_{k \in \mathcal{K}_{\text{student}}} \mathrm{KL}\!\bigl(p_{s,k}, g_k\bigr).
\end{equation}
This term ensures the student learns to localize the SpineTrack keypoints from ground-truth annotations.

\paragraph{Distillation Loss}

To retain the teacher’s performance on the original body keypoints and a more general pretraining data distribution, we introduce a distillation loss for the overlapping set $\mathcal{K}_{\text{body}}$. For each keypoint $k \in \mathcal{K}_{\text{body}}$, the teacher provides a distribution $p_{t,k}$, and the student produces $p_{s,k}$. We define
\begin{equation}
    \mathcal{L}_{\text{distill}} \;=\; \sum_{k \in \mathcal{K}_{\text{body}}} \mathrm{KL}\!\bigl(p_{s,k},\, p_{t,k}\bigr).
\end{equation}
This aligns the student’s predictions with the teacher’s on known body joints, helping the student approximate teacher's distribution.

\paragraph{Structure Loss}
While the positional loss ensures keypoint-level accuracy, we introduce a \emph{structure loss}, $\mathcal{L}_{\text{structure}}$, to enforce anatomical consistency by minimizing a bone orientation penalty.

Let $\mathcal{B}$ be the set of bone segments in the predicted skeleton. For each segment, we compute the orientation difference between the predicted ($\phi^p_b$) and ground-truth ($\phi^g_b$) bone angles relative to the horizontal axis, and sum to get the structure loss:
\begin{align}
    \Delta \phi_b &= \mathrm{atan2}\!\bigl(\sin(\phi^p_b - \phi^g_b),\, \cos(\phi^p_b - \phi^g_b)\bigr), \\
    \mathcal{L}_{\text{structure}} &= \frac{1}{\pi|\mathcal{B}|} \sum_{b \in \mathcal{B}} \bigl|\Delta \phi_b\bigr|.
\end{align}

\paragraph{Spine Smoothness Loss}
Finally, we regularize consecutive vertebral keypoints to form a smooth curve. Let \(\{\mathbf{s}_i\}_{i=1}^{n}\) denote the ordered spine keypoints extracted from the predicted pose, where each \(\mathbf{s}_i \in \mathbb{R}^2\). We define a differentiable smoothing operation as follows. For \(i = 2, \ldots, n\), the smoothed keypoint \(\tilde{\mathbf{s}}_i\) is computed iteratively:
\begin{align}
\tilde{\mathbf{s}}_i &= \tilde{\mathbf{s}}_{i-1} + w_i \left( \mathbf{s}_i - \tilde{\mathbf{s}}_{i-1} \right), \\
w_i &= 1 - 0.5\, \sigma\Big( \alpha \big( \|\mathbf{s}_i - \tilde{\mathbf{s}}_{i-1}\| - T \big) \Big),
\end{align}
where \(\sigma(\cdot)\) denotes the sigmoid function, \(T\) is a distance threshold, and \(\alpha\) controls the sharpness of the transition. When the distance \(\|\mathbf{s}_i - \tilde{\mathbf{s}}_{i-1}\|\) is below \(T\), \(\sigma\) approximates 0 and \(w_i \approx 1\), meaning little or no adjustment is made. Conversely, when the distance exceeds \(T\), \(w_i\) approaches 0.5, effectively dampening abrupt changes.

The \emph{spine smoothness loss} is then defined as the mean squared error between $\mathbf{s}_i$ and $\tilde{\mathbf{s}}_i$:
\begin{equation}
\mathcal{L}_{\text{spine}} = \operatorname{MSE}\Big(\{\mathbf{s}_i\}, \{\tilde{\mathbf{s}}_i\}\Big).
\end{equation}
This loss encourages the predicted spine to follow a smooth trajectory, consistent with anatomical constraints.

The \textbf{overall minimization objective} is given by a weighted sum of all four losses:

\begin{equation}
\label{eq:final_loss}
\mathcal{L}_{\text{total}} \;=\; \alpha\,\mathcal{L}_{\text{pos}}
\;+\; \beta\,\mathcal{L}_{\text{distill}}
\;+\; \gamma_1\,\mathcal{L}_{\text{structure}}
\;+\; \gamma_2\,\mathcal{L}_{\text{spine}},
\end{equation}
where $\alpha,\beta,\gamma_1,\gamma_2$ are scalar weights. By tuning these weights, we balance the primary goal of keypoint accuracy, the retention of teacher knowledge, and anatomically consistent poses. This integrated framework allows the student to learn an extended skeleton without sacrificing the robust generalization acquired by the teacher model.
\begin{table*}[ht]
  \centering
  \scriptsize
  \caption{\textbf{Performance on Benchmark Datasets.} We compare our top-down SpinePose approach (\textcolor{gray!85}{gray}), which uses a coordinate classification head introduced in~\cite{jiang2023rtmpose}, with all existing methods belonging to the same architectural family. Primary evaluations on our novel SpineTrack dataset show our model outperforming baselines in all cases, while results on large-scale in-the-wild datasets (COCO~\cite{lin2014microsoft} and Halpe26~\cite{alphapose}) demonstrate the effectiveness of our framework in retaining the teacher's performance. In particular, unlike other techniques trained over several hundred epochs using combinations of multiple datasets, our approach involves fine-tuning a Body8-pretrained model for just 10 epochs. The methods are grouped according to model size. \textbf{Bold} and \underline{underlined} values denote the best and second-best performance within each group, respectively. All results provided use the flip test. For the COCO and Halpe26 datasets, a detector with 56.4 AP is used, while GT bounding boxes are used for SpineTrack. Models marked with an asterisk (*) have an input size of 384x288, whereas all other models use an input size of 256x192.}
  \label{tab:results}
  \begin{tabularx}{\linewidth}{llXYYYYYYYYYYYYcc}
    \toprule
    & & & \multicolumn{2}{c}{COCO} & \multicolumn{2}{c}{Halpe26} & \multicolumn{2}{c}{Body} & \multicolumn{2}{c}{Feet} & \multicolumn{2}{c}{Spine} & \multicolumn{2}{c}{Overall} & & \\
    \cmidrule(lr){4-5}
    \cmidrule(lr){6-7}
    \cmidrule(lr){8-9}
    \cmidrule(lr){10-11}
    \cmidrule(lr){12-13}
    \cmidrule(lr){14-15}
    Method & Train Data  & Kpts  & AP & AR & AP & AR & AP & AR & AP & AR & AP & AR & AP & AR & Params (M) & FLOPs (G)\\
    
    \cmidrule{1-3}
    \cmidrule(lr){4-7}
    \cmidrule(lr){8-15}
    \cmidrule(lr){16-17}

    SimCC-MBV2     & COCO       &  17 & 62.0 & 67.8 & 33.2 & 43.9 & 72.1 & 75.6 &  0.0 &  0.0 &  0.0 &  0.0 &  0.0 &  0.1 &  2.29 &  0.31 \\
    RTMPose-t      & Body8      &  26 & 65.9 & 71.3 & 68.0 & 73.2 & 76.9 & 80.0 & 74.1 & 79.7 &  0.0 &  0.0 & 15.8 & 17.9 &  3.51 &  0.37 \\
    
    RTMPose-s      & Body8      & 26 & \textbf{69.7} & \textbf{74.7} & \textbf{72.0} & \textbf{76.7} & \textbf{80.9} & \textbf{83.6} & \textbf{78.9} & \textbf{83.5} &  0.0 &  0.0 & 17.2 & 19.4 &  5.70 &  0.70 \\

    \rowcolor{gray!25}
    SpinePose-s    & SpineTrack & 37 & \underline{68.2} & \underline{73.1} & \underline{70.6} & \underline{75.2} & \underline{79.1} & \underline{82.1} & \underline{77.5} & \underline{82.9} & \textbf{89.6} & \textbf{90.7} & \textbf{84.2} & \textbf{86.2} &  5.98 &  0.72 \\
    
    \cmidrule{1-3}
    \cmidrule(lr){4-7}
    \cmidrule(lr){8-15}
    \cmidrule(lr){16-17}
    
    SimCC-ViPNAS   & COCO       & 17 & 69.5 & 75.5 & 36.9 & 49.7 & 79.6 & 83.0 &  0.0 &  0.0 &  0.0 &  0.0 &  0.0 &  0.2 &  8.65 &  0.80 \\
    RTMPose-m      & Body8      & 26 & \textbf{75.1} & \textbf{80.0} & \textbf{76.7} & \textbf{81.3} & \textbf{85.5} & \textbf{87.9} & \textbf{84.1} & \textbf{88.2} &  0.0 &  0.0 & 19.4 & 21.4 & 13.93 &  1.95 \\
    \rowcolor{gray!25}
    SpinePose-m    & SpineTrack & 37 & 73.0 & 77.5 & 75.0 & 79.2 & 84.0 & 86.4 & 83.5 & 87.4 & \textbf{91.4} & \textbf{92.5} & \textbf{88.0} & \textbf{89.5} & 14.34 &  1.98 \\
    
    \cmidrule{1-3}
    \cmidrule(lr){4-7}
    \cmidrule(lr){8-15}
    \cmidrule(lr){16-17}

    RTMPose-l      & Body8      &  26 & \textbf{76.9} & \textbf{81.5} & \textbf{78.4} & \textbf{82.9} & \textbf{86.8} & \textbf{89.2} & \textbf{86.9} & \textbf{90.0} &  0.0 &  0.0 & 20.0 & 22.0 & 28.11 &  4.19 \\
    RTMW-m         & Cocktail14 & 133 & 73.8 & 78.7 & 63.8 & 68.5 & 84.3 & 86.7 & 83.0 & 87.2 &  0.0 &  0.0 &  6.2 &  7.6 & 32.26 &  4.31 \\
    SimCC-ResNet50 & COCO       &  17 & 72.1 & 78.2 & 38.7 & 51.6 & 81.8 & 85.2 &  0.0 &  0.0 &  0.0 &  0.0 &  0.0 &  0.2 & 36.75 &  5.50 \\
    \rowcolor{gray!25}
    SpinePose-l    & SpineTrack &  37 & \underline{75.2} & \underline{79.5} & \underline{77.0} & \underline{81.1} & \underline{85.4} & \underline{87.7} & \underline{85.5} & \underline{89.2} & \textbf{91.0} & \textbf{92.2} & \textbf{88.4} & \textbf{90.0} & 28.66 &  4.22 \\
    
    \cmidrule{1-3}
    \cmidrule(lr){4-7}
    \cmidrule(lr){8-15}
    \cmidrule(lr){16-17}

    SimCC-ResNet50*& COCO       &  17 & 73.4 & 79.0 & 39.8 & 52.4 & 83.2 & 86.2 &  0.0 &  0.0 &  0.0 &  0.0 &  0.0 &  0.3 & 43.29 & 12.42 \\
    RTMPose-x*     & Body8      &  26 & \textbf{78.8} & \textbf{83.4} & \textbf{80.0} & \textbf{84.4} & \textbf{88.6} & \textbf{90.6} & \textbf{88.4} & \textbf{91.4} &  0.0 &  0.0 & 21.0 & 22.9 & 50.00 & 17.29 \\

    RTMW-l*        & Cocktail14 & 133 & 75.6 & 80.4 & 65.4 & 70.1 & 86.0 & 88.3 & 85.6 & 89.2 &  0.0 &  0.0 &  6.5 &  8.1 & 57.20 &  7.91 \\

    RTMW-l*        & Cocktail14 & 133 & \underline{77.2} & \underline{82.3} & 66.6 & 71.8 & \underline{87.3} & \underline{89.9} & \underline{88.3} & \underline{91.3} &  0.0 &  0.0 &  6.9 &  8.6 & 57.35 & 17.69 \\

    \rowcolor{gray!25}
    SpinePose-x*   & SpineTrack &  37 & 75.9 & 80.1 & \underline{77.6} & \underline{81.8} & 86.3 & 88.5 & 86.3 & 89.7 & \textbf{89.3} & \textbf{91.0} & \textbf{88.3} & \textbf{89.9} & 50.69 & 17.37 \\
    \bottomrule
  \end{tabularx}
\end{table*}

\section{Experiments}
\label{sec:experiments}

This section outlines the experiments conducted to evaluate our SpinePose approach on public human pose estimation benchmarks as well as our SpineTrack dataset. We describe the baselines, metrics, and training protocol used, then present our results.

\paragraph{Baselines}
We compare SpinePose against multiple variants of RTMPose~\cite{jiang2023rtmpose} trained on well-known datasets such as COCO~\cite{lin2014microsoft}, Halpe26~\cite{alphapose}, and Body8\footnote{Body8 is an internal merge from~\cite{jiang2023rtmpose} which includes COCO, MPII, CrowdPose, Halpe, PoseTrack18, JHMDB, AIC, and OCHuman datasets}. The table groups methods by model size, allowing us to examine how extending the RTMPose skeleton with our spine keypoints affects performance across model scales.

\paragraph{Evaluation Metrics}
We follow standard COCO-style keypoint evaluation, reporting average precision (AP) and average recall (AR) across varying overlap thresholds. For SpineTrack, we subdivide keypoints into \emph{body}, \emph{feet}, and \emph{spine} subsets, and additionally provide an \emph{overall} AP/AR across all 37 keypoints. Ground-truth bounding boxes are used for SpineTrack, while a pretrained person detector is used for COCO and Halpe26.

\paragraph{Training Protocol}
All SpinePose models begin from Body8-pretrained RTMPose weights. We fine-tune for \(\mathtt{max\_epochs}=10\), using AdamW~\cite{loshchilov2017decoupled} with an initial learning rate of \(4\times10^{-3}\). A short linear warm-up runs for the first 1000 iterations, after which a cosine annealing schedule decays the learning rate to \(5\%\) of its initial value by the end of training. We set the batch size to fit within GPU memory, scaling linearly to maintain a consistent effective batch size of 1024 across experiments. Unless otherwise noted, all models are trained and evaluated using an input size of \(256\times192\) for a fair comparison.  

\paragraph{Results}

Table~\ref{tab:results} highlights the effectiveness of our SpinePose approach compared to existing top-down methods of similar architecture and scale. Notably, our models—fine-tuned for just 10 epochs on the newly introduced spine keypoints—maintain strong performance on large-scale in-the-wild datasets (COCO and Halpe26) while achieving superior results on SpineTrack. This outcome confirms that extending the skeleton to include detailed spinal landmarks does not compromise the original body joints; rather, it can enhance the model’s overall representation of human pose.

\begin{figure*}[ht]
    \centering
    \includegraphics[width=\textwidth]{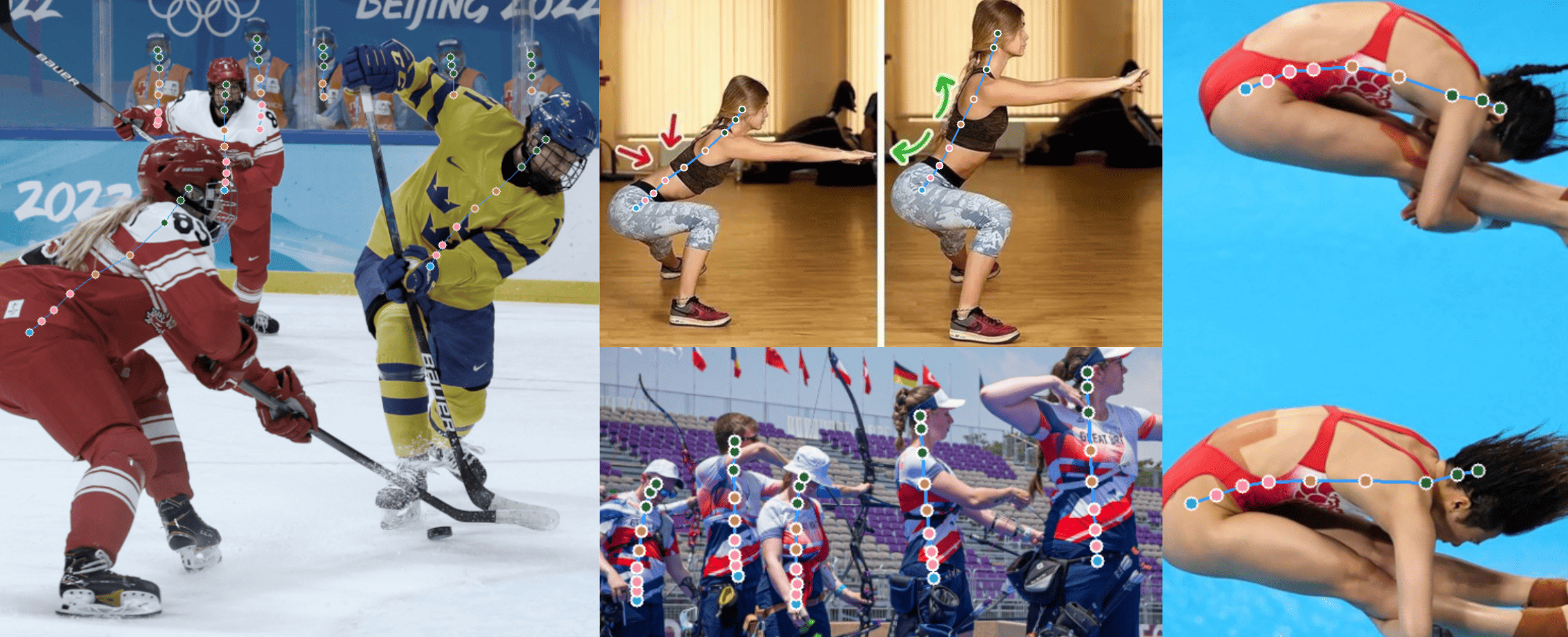}
    \caption{\textbf{Qualitative Results.} Our SpinePose model accurately localizes extended spine landmarks in a variety of sporting activities, including ice hockey, weightlifting, and diving, as well as group training scenarios. The new spine keypoints (shown in pink-to-orange hues) remain stable under heavy occlusions and extreme poses, aligning well with the existing body keypoints (e.g., shoulders, hips, feet). This enriched anatomical detail fosters better posture analysis in sports and potential applications in healthcare and rehabilitation. We omit non-spine keypoints for a cleaner visualization. Further qualitative results are provided in the supplementary materials.}
    \label{fig:results_qualitative}
\end{figure*}

\paragraph{Qualitative Discussion}
Figure~\ref{fig:results_qualitative} shows spine predictions across a diverse range of athletic and fitness scenarios, reinforcing the quantitative findings discussed later in the ablation study. The left panel illustrates two hockey players engaged in a fast-paced tackle, where our model effectively tracks the torsos despite heavy occlusions and non-traditional poses. In the central images, the spine alignment of a squatting athlete remains stable throughout the motion, helping to reveal subtle posture deviations that can be crucial for injury prevention and performance optimization. On the right, the model accurately locates the spine curvature during a mid-air diving sequence, capturing pronounced trunk flexion. The group archery scene similarly demonstrates consistent trunk keypoint placement, even with high crowd density and overlapping limbs. These results emphasize the value of including spine points for analyzing sports movements that rely heavily on core stability and posture. By explicitly modeling the spinal column, our approach detects key torsional and flexion angles integral to sports science, rehabilitation, and coaching use cases. Beyond sports, this broader skeletal representation also opens doors for future research in clinical assessments where spine alignment is a critical indicator of musculoskeletal health. This is a significant advancement over current state-of-the-art methods that represent the spine with a straight line, and can be used as a stepping stone towards clinical-grade spine pose estimation in uncontrolled settings.

\begin{table}[t]
  \centering
  \scriptsize
  \caption{\textbf{Ablations.} We study the impact of our distillation-based training and specialized loss components using SpinePose-l, showing how each contributes to the performance of our system. All models use a Body8-pretrained teacher and train the student for 10 epochs, with $\beta=2.5$, $\gamma_1=0.1$ and $\gamma_2=0.5$ when $\mathcal{L}_\mathrm{distill}$, $\mathcal{L}_\mathrm{structure}$, and $\mathcal{L}_\mathrm{spine}$ are enabled (\checkmark), respectively, and set to zero when disabled. $\alpha=5$ is fixed for the keypoint position loss.}
  \label{tab:ablations}
  \begin{tabularx}{\linewidth}{cccYYY}
    \toprule
    $\mathcal{L}_\mathrm{distill}$ & $\mathcal{L}_\mathrm{structure}$ & $\mathcal{L}_\mathrm{spine}$ & COCO AP & Halpe26 AP & SpineTrack AP \\ 
    \midrule
               &            &            & 69.6 & 72.2 & 87.0 \\
    \checkmark &            &            & 70.8 & 73.2 & 87.3 \\
    \checkmark &            & \checkmark & \underline{74.7} & \underline{76.6} & \textbf{88.7} \\
    \checkmark & \checkmark &            & 71.2 & 73.4 & 87.7 \\
    \rowcolor{gray!25}
    \checkmark & \checkmark & \checkmark & \textbf{75.2} & \textbf{77.0} & \underline{88.4} \\
    \bottomrule
  \end{tabularx}
\end{table}

\noindent \textbf{Ablations} \quad We conduct an ablation study to evaluate the impact of each loss component in our framework using the SpinePose-l model. The results are shown in Table~\ref{tab:ablations}.

Row 1 represents the baseline where the student model (initialized with teacher weights) is trained on the SpineTrack dataset without additional supervision from the teacher. This can be seen as fine-tuning a pretrained model on SpineTrack with no specialized mechanisms for retaining generalization on pretraining datasets or boosting learning. As expected, while the model achieves 87.0 AP on the SpineTrack validation set, its AP on the benchmark in-the-wild datasets is the lowest among all tested combinations. This is remedied by adding a distillation loss between the teacher and the student (row 2), with a 1.0-1.2 point increase on validation sets of pretraining datasets. This indicates that the distillation loss prevents the student from diverging too far from the teacher's data distribution, as intended. Interestingly, the added regularization also results in a slight improvement (+0.3 points) on the SpineTrack validation set. In row 3, we add the spine regularization term, providing the student a strong signal about the anatomy of the new keypoints in the extended skeleton. This results in a significant jump (+1.4 points) on the SpineTrack validation set, indicating the model could better learn the spine anatomy without overfitting on the training samples. This anatomical regularization also benefits performance on pretraining datasets where performance improves even more drastically (+3.9 and +3.4 respectively). This can be explained by the rule-based knowledge the model can learn about the newly added keypoints, which prevents it from interfering with positional knowledge previously learned for the base skeleton. When used independently (row 4), structure loss also leads to similar behavior as the spine loss, but on a smaller scale. We observe performance increases in 0.2-0.4 point range on all three datasets. We achieve highest performance gains over the baseline by combining the structure loss and spine smoothness loss into our integrated pipeline with both distillation and keypoint losses (row 5). These results confirm the design choices made in our SpinePose architecture, clearly demonstrating the impact of each component.

\section{Conclusion}
\label{sec:conclusions}

We presented SpineTrack, a comprehensive dataset tailored for 2D spine pose estimation in unconstrained settings, combining both real-world and synthetic imagery. Our active learning annotation pipeline—coupled with biomechanical validation—ensures anatomically consistent labels that capture the nuanced curvature of the spinal column. Through the proposed SpinePose framework, we demonstrated how detailed vertebral keypoints can be seamlessly integrated into existing pose estimation pipelines without sacrificing overall accuracy. This opens the door to more precise analyses of posture and movement in sports, where spinal alignment is often critical to performance and injury prevention.

Beyond validating the feasibility of high-resolution spine tracking, our work sets a foundation for future research in biomechanics, healthcare, and sports science. Possible extensions include incorporating 3D annotations to better capture depth-aware spinal motion and exploring semi-supervised or self-supervised strategies to reduce manual labeling requirements. The synergy between anatomically rich datasets like SpineTrack and cutting-edge models offers new opportunities for real-time monitoring and technique optimization in athletic training. We believe these advances will further solidify the role of fine-grained spine analysis in driving innovation across sports performance assessment and beyond.

\section*{Acknowledgment}

The work leading to this publication was partially funded by the European Union’s Horizon Europe research and innovation programme under grant agreement No. 10192889.

{\small
\bibliographystyle{ieee_fullname}
\bibliography{main}
}


\end{document}